%% file: main.tex
\documentclass[10pt,twocolumn,letterpaper]{article}

\usepackage{iccv}
\usepackage{times}
\usepackage{epsfig}
\usepackage{graphicx}
\usepackage{amsmath}
\usepackage{amssymb}

\usepackage{graphicx}
\usepackage{amsmath}
\usepackage{amssymb}
\usepackage{booktabs}
\usepackage{soul}

\usepackage{tabularx}
\usepackage{float}
\usepackage{array}
\usepackage{mwe}
\usepackage{subcaption}
\usepackage{mathtools}

\usepackage[pagebackref=true,breaklinks=true,colorlinks,bookmarks=false]{hyperref}
\usepackage[accsupp]{axessibility}

\usepackage[capitalize]{cleveref}
\crefname{section}{Sec.}{Secs.}
\Crefname{section}{Section}{Sections}
\Crefname{table}{Table}{Tables}
\crefname{table}{Tab.}{Tabs.}

\iccvfinalcopy %

\ificcvfinal\fi

\input{preamble}
\begin{document}

\definecolor{turquoise}{cmyk}{0.65,0,0.1,0.3}
\definecolor{purple}{rgb}{0.65,0,0.65}
\definecolor{dark_green}{rgb}{0, 0.5, 0}
\definecolor{orange}{rgb}{0.8, 0.6, 0.2}
\definecolor{red}{rgb}{0.8, 0.2, 0.2}
\definecolor{darkred}{rgb}{0.6, 0.1, 0.05}
\definecolor{blueish}{rgb}{0.0, 0.3, .6}
\definecolor{light_gray}{rgb}{0.7, 0.7, .7}
\definecolor{pink}{rgb}{1, 0, 1}
\definecolor{greyblue}{rgb}{0.25, 0.25, 1}

\title{Generalizing Neural Human Fitting to Unseen Poses With\\ Articulated SE(3) Equivariance}

\author{
Haiwen Feng\affmark[1] \quad
Peter Kulits\affmark[1] \quad
Shichen Liu\affmark[2] \quad
Michael J. Black\affmark[1] \quad
Victoria Abrevaya\affmark[1]\\
\affaddr{\affmark[1]Max Planck Institute for Intelligent Systems, T\"ubingen, Germany} \\ 
\affaddr{\affmark[2]University of Southern California}\\
\email{\{hfeng,kulits,black,vabrevaya\}@tuebingen.mpg.de},\quad \email{\{liushich\}@usc.edu}
}

\twocolumn[{
\renewcommand\twocolumn[1][]{#1}%
\maketitle
\begin{center}
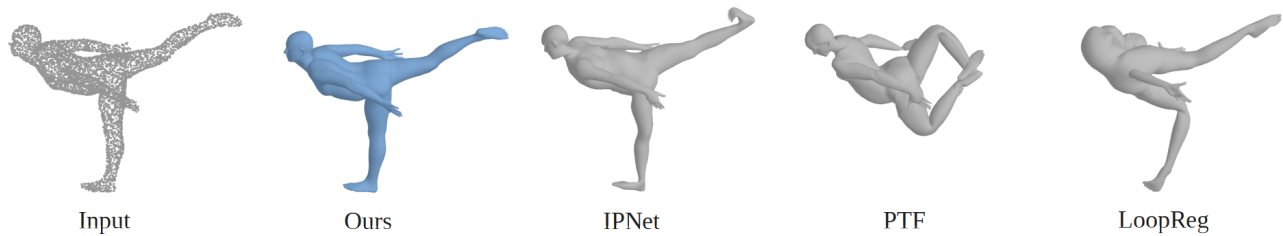

\vspace{-0.9cm}
\input{figures/tex/teaser}
\vspace{-0.24in}
\captionof{figure}{
\modelname is a part-based SE(3)- invariant/equivariant architecture that estimates SMPL \cite{Loper15} body parameters from point clouds. We show results (Ours) on an out-of-distribution pose, compared with state-of-the-art methods  IP-Net~\cite{Bhatnagar20a}, PTF~\cite{Wang21}, and LoopReg~\cite{Bhatnagar20b}. Note that IP-Net's output has a flipped torso. 
}
\label{fig:teaser}
\end{center}
}]
\ificcvfinal\thispagestyle{empty}\fi
\input{sections/0_Abstract.tex}

\input{sections/1_Introduction.tex}
\input{sections/2_Related.tex}

\input{sections/3_Method2.tex}

\input{sections/4_Results.tex}

\input{sections/5_Conclusions.tex}

{\small
\bibliographystyle{ieee_fullname}
\bibliography{references}
}

\clearpage
\appendix
{\noindent\LARGE\textbf{Supplementary Material}}
\newline
\renewcommand{\thefigure}{S.\arabic{figure}}
\renewcommand{\thetable}{S.\arabic{table}}
\renewcommand{\theequation}{S.\arabic{equation}}
\setcounter{figure}{0}
\setcounter{table}{0}
\setcounter{equation}{0}
\input{sections/suppmat/results}
\input{sections/suppmat/implementation}

\input{sections/suppmat/limitations}
\end{document}

%% file: preamble.tex
\usepackage{paralist}   %

\usepackage[numbers,sort,compress]{natbib}

\newcommand{\modelname}{ArtEq\xspace}

\newcommand{\real}{\mathbb{R}}

\renewcommand{\paragraph}[1]{\vspace{1em}\noindent\textbf{#1}.}

\newcommand{\eqfeat}{\mathcal{F}}
\newcommand{\partfeat}{\mathcal{H}}
\newcommand{\partk}{\vect{p}_k}
\newcommand{\point}{\vect{x}_i}
\newcommand{\group}{\mathcal{G}}
\newcommand{\rotation}{\mathcal{R}}

\newcommand{\inputpc}{\mathbf{X}}

\newcommand{\vect}[1]{\mathbf{#1}}

\newcommand{\shapecoeff}{\boldsymbol{\beta}}
\newcommand{\shapedim}{{\left| \shapecoeff \right|}}

\newcommand{\posecoeff}{\boldsymbol{\theta}}

\newcommand{\template}{\mathbf{\bar{T}}}
\newcommand{\restpose}{\mathbf{T}}

\newcommand{\blendweights}{\mathcal{W}}

\DeclareMathOperator*{\argmin}{arg\,min}

\newcommand*{\affaddr}[1]{#1} 
\newcommand*{\affmark}[1][*]{\textsuperscript{#1}}
\newcommand*{\email}[1]{\small{\texttt{#1}}}

%% file: figures/tex/teaser.tex
\includegraphics[width=\linewidth]{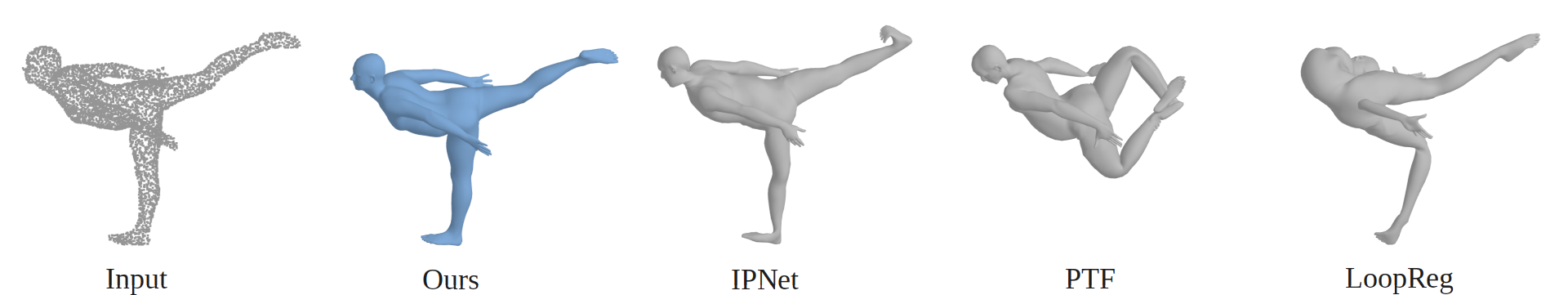}

%% file: sections/0_Abstract.tex
\begin{abstract}
We address the problem of fitting a parametric human body model (SMPL) to point cloud data. 
Optimization-based methods require careful initialization and are prone to becoming trapped in local optima.
Learning-based methods address this but %
do not generalize well when the input pose is far from those seen during training. 
For rigid point clouds, remarkable generalization has been achieved by leveraging SE(3)-equivariant networks, but these methods do not work on articulated objects. 
In this work we extend this idea to human bodies and propose \modelname, a novel part-based SE(3)-equivariant neural architecture for SMPL model estimation from point clouds. 
Specifically, we learn a part detection network by leveraging local SO(3) invariance, and regress shape and pose using articulated SE(3) shape-invariant and pose-equivariant networks, all trained end-to-end. 
Our novel pose regression module leverages the permutation-equivariant
property of self-attention layers to preserve rotational equivariance.
Experimental results show that \modelname generalizes to poses not seen during training, outperforming state-of-the-art methods by $\sim$44\% in terms of body reconstruction accuracy, without requiring an optimization refinement step. Furthermore, \modelname
is three orders of magnitude faster during inference than prior work and has 
97.3\% fewer parameters.
The code and model are available for research purposes at \url{https://arteq.is.tue.mpg.de}.
\vspace*{-0.5cm}
\end{abstract}

%% file: sections/1_Introduction.tex
\section{Introduction}
\label{sec:intro}
The three-dimensional (3D) capture of humans in varied poses is increasingly common and has many applications including synthetic data generation~\cite{Patel:CVPR:2021}, human health analysis~\cite{WONG2023}, apparel design and sizing~\cite{tsoliWACV14}, and avatar creation~\cite{Chen21,Peng21,Tiwari21,Wang22a}.
Existing 3D body scanners output unordered point clouds, which are not immediately useful for the above applications.
Consequently, the first step in processing such data is to \emph{register} it; that is, to transform it 
into a canonical and consistent 3D representation such as a mesh with a fixed topology.
For human bodies, this is typically done by first fitting a parametric model like SMPL~\cite{Loper15} to the data; see Figure  \ref{fig:teaser}.
Such a process should be efficient and general; that is, it should work for any input body scan, no matter the complexity of the pose. 
However, this is challenging given the articulated structure of the body and the high degree of variation in shape and pose.

Traditional optimization-based methods for fitting bodies to point clouds~\cite{Allen03, Bogo14, Bogo17, Joo18, Hirshberg:ECCV:2012} are usually based on ICP~\cite{Chen92} or its variants~\cite{Allen03, PonsMoll15, Zuffi15}. 
These approaches can recover accurate results even for complex poses, but  
require a good initialization, are computationally expensive, and may require significant manual input.
Inspired by progress made in neural architectures for rigid 3D point clouds~\cite{Qi17, Qi17b, Zaheer17, Thomas19},
learning-based approaches have been proposed to solve the registration task. 
Previous approaches directly regress model parameters~\cite{Jiang19, Wang21, Liu21}, 
intermediate representations such as correspondences~\cite{Bhatnagar20a, Bhatnagar20b}, or %
mesh vertex positions~\cite{Groueix18, Prokudin19, Zhou20}. 
While less accurate than optimization, they can be used to initialize an optimization-based refinement step for improved accuracy~\cite{10.1111:cgf.14502}.

A major limitation of learning-based approaches, as reported in several recent papers~\cite{Bhatnagar20a, Bhatnagar20b, Wang21}, is their \textbf{poor generalization to body poses that lie outside the training distribution.}
To understand why, let us first consider how a parametric model such as SMPL explains the input point cloud. Given the shape parameters, the model first generates the overall body structure
by deforming a template mesh in a canonical pose. Pose-dependent offsets are then added to the mesh.
This deformed mesh then undergoes an articulated transformation that poses
the body parts rigidly, and %
linear blend skinning is applied to smooth the result.
Therefore, the observed point cloud is modeled as a combination of a canonical body shape, a part-based articulated model, and non-rigid pose-corrective deformations. 
When training networks to fit SMPL to point clouds, 
the networks are tasked with capturing the joint distribution of canonical shape and pose deformation, entangling these factors while learning a prior over plausible body shape and pose. 
This data-dependent prior is useful to infer new, in-distribution samples, but becomes a limitation when it comes to poses that are far from the training set.  
Ideally, if the networks were designed to be \emph{equivariant} to articulated body pose transformations, then unseen body poses at test time would not be a problem.

A function (network) 
$f : V \to W$ is said to be equivariant with respect to %
a group $\group$
if, for any transformation $\mathcal{T} \in \group, f(\mathcal{T} \vect{X}) = \mathcal{T} f(\vect{X})$, $\vect{X} \in V$.
This property 
can aid in generalization
since it allows one to train with only ``canonical'' inputs $f(\vect{X})$, while generalizing, by design, to any transformation of the group $\mathcal{T} f(\vect{X})$. For example, SE(3)-equivariant networks have been used to address the out-of-distribution (OOD) generalization problem in rigid point cloud tasks, see for example \cite{Deng21, Puny21, Chen21b}.
However, extending this to the human body is far from straightforward, due to 1) its high degree of articulation, 2) the entanglement of pose and shape, and 3) deformations that are only approximately rigid.

In this work we introduce \modelname, a new neural method that regresses SMPL shape and pose parameters from a point cloud.
Our key insight is that non-rigid deformations of the human body can be largely approximated as part-based, articulated, rigid SE(3) transformations, and that good generalization requires 
a proper integration of equi-/in-variant properties into the network.
With this in mind, we propose a novel equi-/in-variant architecture design based on the discretized SE(3)-equivariant framework~\cite{Chen21b, Cohen19}. %
We learn a part detection network by leveraging local SO(3) invariance
and regress shape and pose by proposing \emph{articulated} SE(3) shape-invariant and pose-equivariant networks, all trained in an end-to-end manner.
We further propose a novel %
pose regression module that leverages the permutation-equivariant property of self-attention layers to preserve rotational equivariance. Finally, to facilitate generalization to unseen poses, we cast pose regression as a weight prediction task, in which the predicted weights are used to calculate a weighted average over each of the discretized SO(3) rotations to obtain the final result.

Our empirical studies demonstrate the importance of introducing SE(3) equi-/in-variance for the task of SMPL pose and shape estimation from point cloud data. For out-of-distribution data, we show significant improvement over competing methods~\cite{Bhatnagar20a, Bhatnagar20b, Wang21} in terms of part segmentation, as well as accuracy in pose and shape estimation, even when 
others 
are trained with SO(3) data augmentation. Notably, we outperform methods that require an optimization step, while \modelname is purely regression-based. 
Our method also shows strong performance for in-distribution samples, surpassing all previous (optimization-based) methods although it only uses regression.
Finally, we demonstrate how employing the right symmetries can lead to a lightweight network that is more than thirty times smaller than prior models, as well as a thousand times faster at inference time, making it easy to deploy in real-world scenarios.

In summary, we make the following contributions:
(1) We propose a new framework for human shape and pose estimation from point clouds that integrates SE(3)-equi-/in-variant properties into the network architecture. 
(2) We propose a novel SE(3)-equivariant pose regression module that combines SE(3) discretization with the permutation-equivariant property of self-attention layers.
(3) We show state-of-the-art performance on common benchmarks and datasets based on only regression, particularly for out-of-distribution poses. Additionally, our framework results in a much lighter model that performs three orders of magnitude faster than competitors at inference time.
Our code and pre-trained models are available at \url{https://arteq.is.tue.mpg.de}.

%% file: sections/2_Related.tex
\section{Related Work}

\paragraph{Human Body Registration From Point Clouds}
Classic approaches for body registration typically deform a template mesh or a parametric model using some variant of the ICP~\cite{Chen92} algorithm~\cite{Allen03, PonsMoll15, Zuffi15}, often with additional cues such as color patterns~\cite{Bogo14, Bogo17} or markers~\cite{Pishchulin17, Allen03, Anguelov05}. These optimization-based methods can produce accurate results for complex poses when properly initialized, but are prone to getting stuck in local minima when not, can be slow to generate results, and are not fully automatic.

Learning-based approaches have gained popularity, largely due to the development of effective neural networks for point cloud processing such as PointNet~\cite{Qi17, Qi17b}, DeepSets~\cite{Zaheer17}, and KPConv~\cite{Thomas19}. These are trained to produce either a good initialization for a subsequent optimization process~\cite{Bhatnagar20a, Groueix18, Wang21} or as end-to-end systems that directly compute either a mesh~\cite{Prokudin19, Wang20, Zhou20} or the parameters of a human body model~\cite{Jiang19, Bhatnagar20b, Zhou20, Liu21}. Our method falls into the last category. 

Model-based registration reduces the task to estimating pose and/or shape parameters of a human body model such as SMPL~\cite{Loper14}. 
It has been noted that it is difficult to directly regress the parameters of a SMPL model from point clouds~\cite{Jiang19, Bhatnagar20a, Bhatnagar20b, Wang20, Wang21}. To circumvent this, current approaches go through intermediate representations such as joint-level features~\cite{Jiang19, Liu21} %
and correspondence maps~\cite{Bhatnagar20b, Wang21}, or resort to temporal data and motion models~\cite{Jiang22}. 
Similar to this work, part-based segmentation has been used as an additional cue for registration~\cite{Li19, Bhatnagar20a, Bhatnagar20b, Wang21}. Closely related to our work, PTF~\cite{Wang21} isolates each segmented part and regresses a local transformation from input space to canonical space, from which pose parameters are obtained via least-squares fitting. %

Without explicitly considering rotational symmetries, 
previous methods struggle to generalize to poses unseen during training, which limits their applicability.
To the best of our knowledge, ours is the first work on model-based human point cloud registration specifically designed to correctly and efficiently handle out-of-distribution poses.

\paragraph{Equivariant Learning on Point Clouds} 
The success of CNNs is largely attributed to the translational equivariance property of such networks. 
Consequently, there has been an increasing interest in making neural networks invariant or equivariant to other symmetry groups~\cite{Zaheer17, Qi17, Keriven19,Sosnovik20,Cohen18, Weiler18, Esteves18,Chen21b, Fuchs20, Deng21,Satorras21, Atzmon22,Cohen16, Puny21}. 
Of particular interest for point cloud processing is the SE(3) equivariance group. Methods for achieving SE(3) equivariance include:  
Vector Neurons~\cite{Deng21}, which employ tensor features and accordingly designed linear/non-linear equivariant layers; 
Tensor Field Networks (TFN)~\cite{Thomas18} and SE3-transformers~\cite{Fuchs20}, which build on top of the SO(3) representation theory with spherical harmonics and Clebsch-Gordan coefficients; 
methods that make use of group averaging theory \cite{Puny21, Atzmon22, Sannai2021EquivariantAI}; 
and methods that employ a discretization of the SO(3) space to achieve equivariance \cite{Cohen19, Chen21b}. 
Within this last group, EPN~\cite{Chen21b} uses separable discrete convolutions (SPConv) that split the 6D SE(3) convolutions into SO(3) and translational parts, improving computational efficiency. 
Here, we employ the discrete SO(3) framework along with SPConv layers, 
as it allows us to leverage 
the discretized space to simplify the problem of pose regression, and has been noted by previous work to be highly effective~\cite{Li21}.

In the case of rigid objects, SE(3)/SO(3)-equivariant networks have been applied to several tasks including classification and retrieval~\cite{Esteves18, Chen21b}, segmentation~\cite{Deng21, Luo22}, registration~\cite{Wang22b, Zhu22}, object manipulation~\cite{Simeonov22}, normal estimation~\cite{Puny21}, and pose estimation/canonicalisation~\cite{Li21, Spezialetti20}.
Also for rigid objects, disentanglement of shape and pose has been considered by some of these works, such as \cite{Katzir22, Li21}. 
To our knowledge, the only method that explores piece-wise equivariance 
for articulated objects is \cite{Puny21}. However, 
this is done in the context of shape-space learning, which takes as input already registered meshes with ground-truth part segmentation information. %
Our work, on the other hand, takes as input unordered and unevenly sampled point clouds, for which the parts are unknown.

%% file: sections/3_Method2.tex
\section{Preliminaries}
\label{sec:preliminaries}

\subsection{Discretized SE(3) Equivariance}
\label{ssec:preliminaries_so3}

\input{figures/tex/discretize}

\input{figures/tex/pipeline}

Gauge-equivariant neural networks were proposed by Cohen et al.~\cite{Cohen19} as a way to extend the idea of 2D convolutions to the manifold domain.
Intuitively, instead of shifting a convolutional kernel through an image for translational equivariance, gauge-equivariant networks ``shift'' a kernel through all possible tangent frames for equivariance to gauge symmetries. %
Since this process is very computationally expensive, \cite{Cohen19} proposed to discretize the SO(3) space with the icosahedron, which is the largest regular polyhedron, exhibiting 60 rotational symmetries. 
This has been further extended to operate on point clouds and the SE(3) space by EPN~\cite{Chen21b}, which introduces separable convolutional layers (SPConv) to independently convolve the rotational and translational parts. 

Formally, the SO(3) space can be discretized by a rotation group $\group$ of size $\lvert \group \rvert = 60$, where each group element $\vect{g}_j$ represents a rotation $\rotation(\vect{g}_j)$ in the icosahedral rotation group.
As shown in \Cref{fig:discretize_rotate}, a rotation acting on a continuous equivariant feature is equivalent to a permutation acting in the discretized space, where the rotation group is permuted in a specific order.
This builds a connection between a rotation in a continuous space and a permutation in a discrete space. 
In this paper, following \cite{Cohen19,Chen21b}, 
we use the rotation group $\group$ and the permutation operator to approximate the SO(3) space and the rotation operator.

\subsection{SMPL Body Model}

SMPL~\cite{Loper15} is a statistical human body model that maps shape $\beta \in \real^{10}$ and pose $\theta \in \real^{K \times 3}$ parameters to mesh vertices $\vect{V} \in \real^{6890 \times 3}$, where $K$ is the number of articulated joints (here $K=22$), and $\theta$ contains the relative rotation of each joint plus the root joint w.r.t.\ the parent in the kinematic tree, in axis-angle representation. The model uses PCA to account for variations in body shape, and Linear Blend Skinning (LBS), $W(\cdot)$, to pose the mesh.

A rest-pose mesh is first produced by
\begin{equation}
    \restpose(\shapecoeff, \posecoeff) = \template + B_S(\shapecoeff) + B_P(\posecoeff), 
\end{equation}
where $\template \in \real^{6890 \times 3}$ is the template mesh, $B_S(\beta)$ is the linear transformation from shape parameters $\shapecoeff$ to shape displacements, and $B_P(\posecoeff)$ is the linear transformation from pose parameters to blendshape correctives that account for soft-tissue deformations due to pose. 
Next, joint locations are obtained for the rest-pose mesh via the linear joint regressor $J(\shapecoeff) : \real^{\shapedim} \to \real^{66}, J(\shapecoeff) = \{ t_1, \dots, t_{22}; t_i \in \real^{3} \} $. 
Finally, SMPL applies LBS using skinning weights $\blendweights$ over the rest-pose mesh to obtain the output vertices:
\begin{equation}
\label{eq:smpl}
    M(\shapecoeff, \posecoeff) = W(
    \restpose(\shapecoeff, \posecoeff), J(\shapecoeff), \posecoeff, \blendweights ). %
\end{equation}

\section{Method}
\label{sec:method}

Given an input point cloud $\inputpc = \{ \vect{x}_i \in \real^3\}_N$ of size $N$ representing a human body, the goal of this work is to regress the SMPL~\cite{Loper15} shape $\shapecoeff$ and pose $\posecoeff$ parameters that best explain the observations
\emph{for any given pose}, including out-of-distribution ones.
Inspired by the progress in rigid SE(3)-equivariant networks~\cite{Atzmon22, Chen21b, Li21}, 
we develop an architecture that extracts part-based SE(3)-  invariant/equivariant features to disentangle shape, pose, and body parts. 

There are key differences between rigid objects and human bodies. 
First, bodies have a highly articulated structure, with different body parts undergoing various SE(3) transformations according to the kinematic tree. Therefore, global SE(3)-equivariant features cannot be directly employed, requiring, instead, local feature extraction and aggregation along with part-based semantic guidance.
Second, most of the human body parts are approximately cylindrical in shape, such as the torso, legs, and arms. Cylindrical shapes have a rotational symmetry, resulting in ambiguities when recovering the pose. This is a known problem in the rigid pose estimation field~\cite{Hodan20}, and cannot be resolved without external information. For human bodies, however, we can use non-ambiguous parts to help disambiguate 
the other ones,
as we will show later. 
Third, when varying body poses the Euclidean neighborhood of a point might not always correspond to the geodesic neighborhood, particularly with poses that are close to self-contact. This is a problem when using point convolutional networks (such as the one developed here) since the kernel has a ball-shaped receptive field that might convolve far-away body points (i.e., close in Euclidean space but far in geodesic space), resulting in incorrect estimates. %

\medskip
To address this
we introduce \textbf{\modelname}, a \emph{part-based SE(3)-equivariant framework for human point clouds} that processes articulated bodies via four modules: (1) shared locally equivariant feature extraction (\cref{ssec:method_local_features}), (2) part segmentation network (\cref{ssec:method_segmentation}), (3) pose estimation network (\cref{sssec:method_pose_estimation}) and (4) shape estimation network (\cref{sssec:method_shape_estimation}). 
An overview of our method can be found in \Cref{fig:pipeline}, and we elaborate each of these in the following.

\subsection{Locally Equivariant Feature Extractor}
\label{ssec:method_local_features}

The first step of our pipeline is to obtain local \emph{per-point} SO(3)-equivariant features that can be used by the subsequent modules.
To this end, we train a network 
that takes as input the point cloud $\inputpc$ and a 
rotation group $\group$ with $\lvert \group \rvert = M$ elements, and returns a feature tensor $\eqfeat \in \real^{N \times M \times C}$, comprising a feature vector of size $C$ for each of the $N$ points and each of the $M$ group elements.

Key to our design is the ability to capture local SO(3)-equivariant features via limiting the point convolution network's effective receptive field.
As mentioned earlier, human body parts that come close to each other are problematic, since the convolutional kernels might incorporate points that are geodesically distant (\eg, an arm that is almost touching a leg). This, in turn, results in reduced performance when body parts are close to each other in Euclidean space. To mitigate this, we employ a small kernel size for the SPConv layer and reduce the number of layers to only two, such that each point-wise feature in $\eqfeat \in \real^{N \times M \times C}$ is calculated using only a small neighboring patch.

\subsection{Part Segmentation via Local SO(3) Invariance}
\label{ssec:method_segmentation}

To obtain part-based equivariance, the first step is to segment the point cloud into body parts.
To achieve this while generalizing to OOD poses, we initially obtain a \emph{locally SO(3)-invariant} feature by average pooling $\eqfeat$ over the rotation group dimension ($M$), aggregating the information from all group elements:
\begin{equation}
    \overline{\eqfeat}(\vect{x}_i) = \sigma\left(\{ \eqfeat(\vect{x}_i, \vect{g}_1), \eqfeat(\vect{x}_i, \vect{g}_2), \dots, \eqfeat(\vect{x}_i, \vect{g}_M) \}\right),
    \label{equ:aggregate}
\end{equation}
where $\sigma$ is an aggregation function (here we use mean pooling), and $\overline{\eqfeat} \in \mathbb{R}^{N \times C}$ 
are the per-point SO(3)-invariant features that encode
intrinsic geometry information.
To efficiently segment the point cloud we adopt a PointNet architecture~\cite{Qi17} with 
skip connections, which takes as input the invariant local feature $\overline{\eqfeat}(\point)$ and outputs a vector $\alpha \in \real^{N \times 22}$, with $\alpha(\point, \partk)$ representing the probability of point $\point$ belonging to body part $\partk$. Note that PointNet is based on set operators such as pooling and point-wise convolution, and hence preserves SO(3) invariance\footnote{In the original PointNet, the input to the first layer is absolute point coordinates which are not invariant to rotations. Here, however, the input is already SO(3)-invariant.}.

\subsection{Pose and Shape Estimation via Attentive SE(3) Equivariance}
\label{ssec:method_pose_shape}

To disentangle pose and shape while maintaining OOD generalization we need to consider the correct symmetries for each task: part-based SE(3) equivariance for pose and part-based SE(3) invariance for shape. We explain here how to process the previously obtained part segmentation and local SO(3)-equivariant features to estimate part-based equivariant features, which are used to %
compute the final SMPL pose $\posecoeff$ and shape $\beta$ parameters.

\subsubsection{Extracting Part-Based Features}
\label{sssec:method_part_features}

The feature tensor $\eqfeat$ operates at the level of points. 
To obtain features that are useful for the part-level task of pose and shape regression, we aggregate $\eqfeat$ into part-based equivariant features $\partfeat \in \real^{K \times M \times C}$.

A na{\"i}ve solution is to simply select the points with maximum probability for a part, and average their features
to obtain a unique equivariant feature of size $M \times C$ for each body part $\partk$.
However, (1) hard selection based on the argmax operation is not differentiable, and (2) points in the transition of two adjacent parts are ambiguous and can be hard to correctly select.
Instead, we propose to use a \emph{soft aggregation} by performing a weighted average of the equivariant features \emph{of all the points}, weighted by the probability computed by the segmentation network:
\begin{equation}
    \partfeat(\vect{p}_k, \vect{g}_j) = \sum_i^N \alpha(\vect{x}_i, \vect{p}_k) \eqfeat(\vect{x}_i, \vect{g}_j),
\end{equation}
where $\mathcal{H} \in \mathbb{R}^{K \times M \times C}$ is a per-part SO(3)-equivariant feature, and $K$ is the number of body parts ($K=22$ in the case of SMPL).
Similar to \Cref{equ:aggregate} in \Cref{ssec:method_segmentation}, we extract the part-level SO(3)-invariant feature by aggregating the equivariant features:
\begin{equation}
    \overline{\mathcal{H}}(\vect{p}_k) = \sigma\left(\{ \mathcal{H}(\vect{p}_k, \vect{g}_1), \mathcal{H}(\vect{p}_k, \vect{g}_2), \dots, \mathcal{H}(\vect{p}_k, \vect{g}_M) \}\right),
\end{equation}
where $\overline{\mathcal{H}}(\vect{p}_k) \in \mathbb{R}^{K \times C}$ is the per-part SO(3)-invariant feature.

\subsubsection{Pose Estimation}
\label{sssec:method_pose_estimation}

\paragraph{Pose Representation} SMPL pose parameters are defined relative to their parent in the kinematic tree. However, local rotations are problematic for equivariant features, since these are defined in a global coordinate system. We estimate instead \emph{global} rotation matrices that represent the rigid transformation from the part in canonical pose to the part in the current pose, from which we can recover the local rotation matrix $\posecoeff_k$ by $\posecoeff_k = \hat{\mathcal{R}}_k \cdot \posecoeff_{parent(k)}^T$, 
where $\hat{\mathcal{R}}_k$ is the estimated global rotation matrix, and $\posecoeff_{parent(k)}$ the accumulated rotation matrix of the parent.%

\paragraph{Attentive SE(3) Equivariance}
To obtain the part rotation matrices $\hat\rotation_k$, we need a function that transforms the part feature vector $\partfeat(\partk)$ into a suitable representation of $\hat\rotation_k$.
This function is required to preserve the equivariance by construction; if it does not (\eg, if we employ a standard MLP), the network must observe the point cloud in all possible poses to be capable of regressing arbitrary rotations.
While we could, in principle, extend the network by concatenating more SPConv layers, this results in larger receptive fields, which are harmful in our particular scenario, and results in longer computation times.
Instead, we can make use of the fact that rotational equivariance in continuous space is equivalent to permutation equivariance in discrete space (see \cite{Chen21b, Cohen16} and \Cref{sec:preliminaries}). Thanks to this, self-attention over group elements is an efficient alternative for preserving rotational equivariance,
and we can use it to extract relationships (\eg, relative importance) among the elements of $\group$.
Hence, we pair our initial SPConv layers with self-attention layers for efficient SE(3)-equivariant processing.

\paragraph{Pose Regression as Group Element Weight Prediction}
Given the set of $M$ group element features, $\partfeat \in \real^{K \times M \times C}$, we now need to regress a rotation for each body part. Here, we can make use of 
the fact that each group element feature is associated with a group element $\vect{g}_j$ (which is a rotation in the discretized SO(3) group -- see \Cref{ssec:preliminaries_so3}).
With this, we can regard the pose regression task as a 
\emph{probabilistic/weighted aggregation of the group element rotations}. Specifically, we use the part-based group element features $\partfeat(\vect{p}_k, \vect{g}_j)$ to regress one weight for each of the $M$ group elements. The final pose is calculated as the weighted chordal L2 mean~\cite{Hartley13} of the group elements with the predicted weights. Since these are predicted by self-attention layers, we can ensure that 1) rotational (permutation) equivariance is preserved and 2) the correlations between the $M$ group element features are being captured. The part-wise rotations can now be regarded as a weighted interpolation between the discrete $M$ group elements, without losing the equivariance of the group element features.

\paragraph{Addressing the Cylindrical Rotational Symmetry} For human bodies, we need to take into account the fact that many parts are of roughly cylindrical shape, resulting in a rotational ambiguity. However, we can leverage the fact that pairs of neighboring parts are less prone to this ambiguity. Imagine an upper leg and lower leg with a bent knee between them. Each body part on its own suffers from rotational ambiguity, but the bent knee means that the complex of both limbs has no such ambiguity.
With this in mind, we condition the pose-estimation network on both a part's feature and the feature of its parent; that is, we concatenate $\partfeat(\partk)$ with the parent feature: $ [ \partfeat(\partk) || \partfeat(\vect{p}_{parent(k)}) ]$, before processing with the self-attention layers. We concatenate the root joint with itself for completeness. 
Experimentally, we find that this helps improve accuracy (\Cref{tab:comparisons}).

\subsubsection{Shape Estimation}
\label{sssec:method_shape_estimation}

Finally, to properly explain the observed point cloud we need to estimate the shape parameters $\shapecoeff$ in a way that is \emph{part-wise-invariant} to pose transformations. To this end, we transform $\partfeat$ into a part-invariant feature by mean pooling over the group element dimension $M$, resulting in a feature matrix $\bar\partfeat \in \real^{K \times C}$, as explained in \Cref{sssec:method_part_features}. This feature is further processed by a few self-attention layers that capture the correlation across different body parts. The output is then flattened and fed into an MLP to produce the final $\shapecoeff$ parameters. 

\subsection{Model Instance and Training}
\label{ssec:method_training}

Thanks to the additional symmetry information, equivariant networks have been shown to be efficient in terms of model size and data requirements~\cite{Cohen19}. We leverage this and instantiate our framework with a minimally sized \modelname~architecture, with only two layers of SPConv that output a feature tensor with channel size $C$ = 64; two multi-head self attention (MHSA) layers for pose regression (eight heads with 64-dim embedding); and one similar MHSA for shape regression. This results in a model that is significantly smaller than competing models~\cite{Wang21}, while still delivering superior performance, as we will see in the next section.

We train the framework in a supervised manner in two stages.
In a first stage, we train the part segmentation network and use ground-truth part segmentation to perform the part-level feature aggregation, while all modules are simultaneously and independently trained.
In the second stage, we use the predictions from the part segmentation network for part-level feature aggregation and train the full pipeline end-to-end.
Training objectives and additional details can be found in the Sup.\ Mat.

%% file: figures/tex/discretize.tex
\begin{figure}[t]
     \centering
     \includegraphics[width=.6\linewidth]{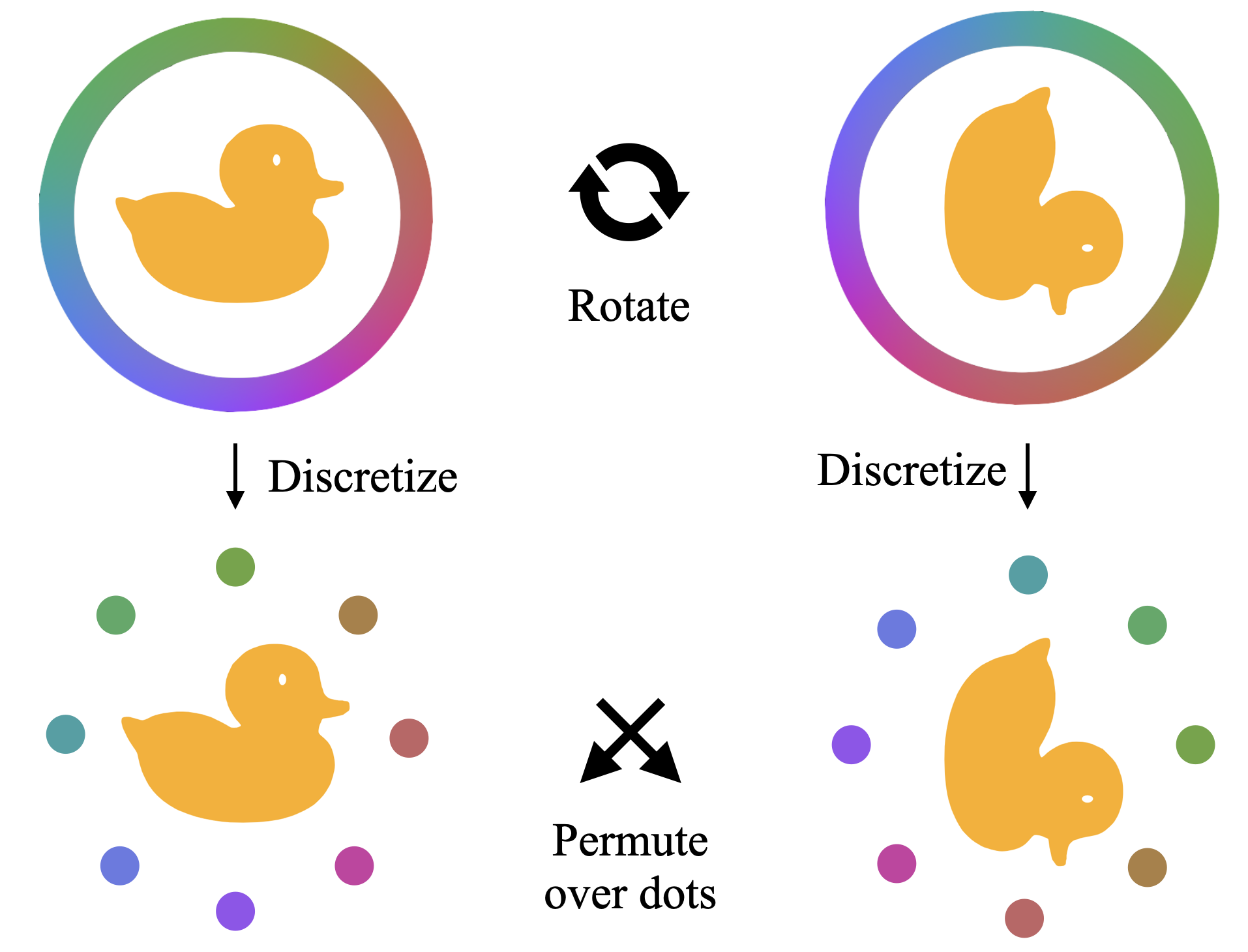}
     
    \caption{A rotation acting in continuous space is equivalent to a permutation acting in the discretized rotation space (in a particular order). We exploit this property to design our architecture based on self-attention layers, while maintaining SO(3) equivariance.}
    \label{fig:discretize_rotate}
\end{figure}

%% file: figures/tex/pipeline.tex
\begin{figure*}[t]
     \centering
     \includegraphics[width=\textwidth]{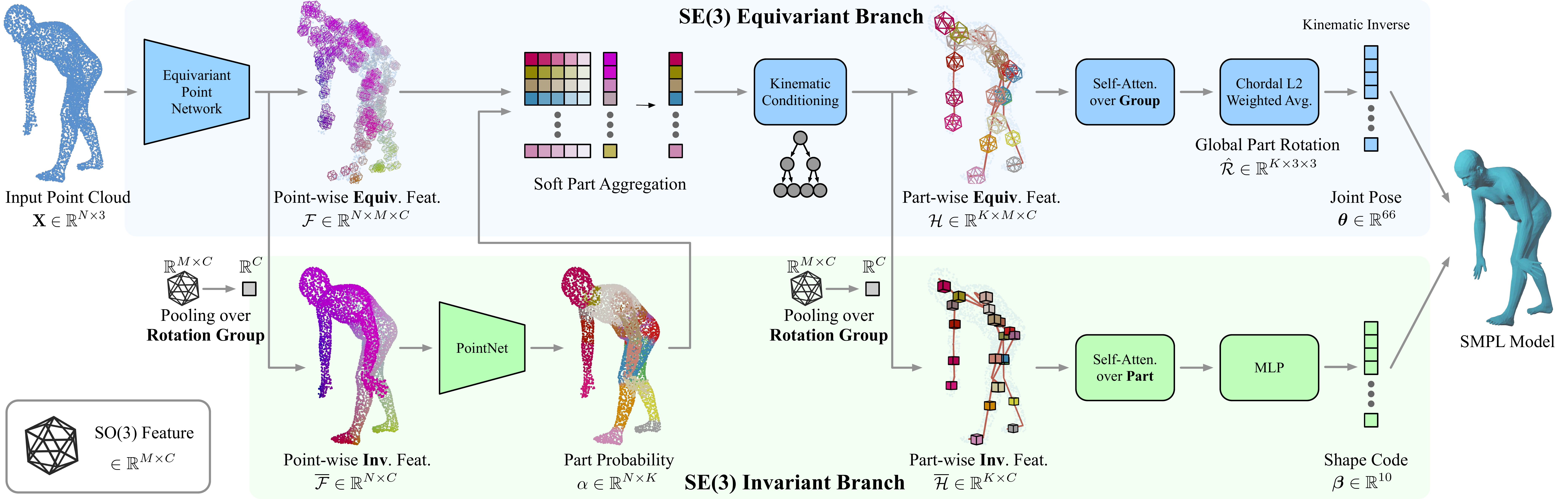}
    \caption{Overview of \textbf{\modelname}. We first obtain point-wise equivariant features using a small equivariant point network \cite{Chen21b}, which provides a $C$-dimensional feature vector per point and per-group element (\ie the $60$ rotational symmetries of the icosahedron). We then convert these into point-wise invariant features by pooling over the rotation group to obtain a part segmentation of the point cloud. Using the segmentation, we softly aggregate the point-wise features into \emph{part-based} equivariant features. A self-attention layer processes these in an efficient manner while preserving equivariance. We cast pose regression as a weight prediction task by predicting the weights necessary to perform a weighted average over each rotation element. Finally, we transform the part-based features into invariant ones to obtain an estimate of the shape.}
    \label{fig:pipeline}
\end{figure*}

%% file: sections/4_Results.tex
\section{Results}
\label{sec:results}

In the following we show qualitative and quantitative results for \modelname, both for in-distribution (ID)
and out-of-distribution (OOD)
data, and we compare with state-of-the-art methods IP-Net~\cite{Bhatnagar20a}, LoopReg~\cite{Bhatnagar20b}, and PTF~\cite{Wang21}. We explain our evaluation protocol in \Cref{sec:results_protocol}, evaluate our SE(3)-invariant segmentation network in \Cref{sec:results_segmentation}, and show quantitative and qualitative performance for SMPL-parameter estimation in \Cref{sec:results_smpl}. Finally, we compare performance time and model size in \Cref{sec:results_performance}. 

\input{tables/tab_segmentation}

\input{figures/tex/segmentation_results}

\subsection{Evaluation Protocol} 
\label{sec:results_protocol}

\noindent\textbf{Datasets.} We train our network using the DFAUST~\cite{Bogo17} subset of the AMASS dataset~\cite{AMASS:ICCV:2019}, which contains 100 sequences of 10 subjects with diverse body shapes. We follow the train-test split used in \cite{Chen21, Atzmon22} and we crop the test sequences to the middle 50\% of frames, subsampling every 5 frames. We sample 5000 non-uniform points across the surface with random on-surface displacements. %
For ID testing we use the test split of DFAUST~\cite{Bogo17}. For OOD testing we use the PosePrior subset~\cite{Akhter15} of AMASS, which contains challenging poses that are far from those performed in DFAUST. 

\vspace{0.02in}
\noindent\textbf{Metrics.} 
The focus of this work is robustness to OOD poses, and hence we employ metrics previously proposed for this goal~\cite{Chen21}. Specifically, for SMPL estimation we measure (1) vertex-to-vertex error (V2V) in cm and (2) joint position error (MPJPE) in cm, which are standard in the human pose community 
and capture  the effect of errors in rotation on the final body, 
propagated through the kinematic tree 
while taking into account the predicted shape.
For body part correspondence estimation, we test accuracy of part segmentation as the percentage of correct assignments.
Note that a high V2V error and MPJPE, or a low part segmentation accuracy suggests a lack of robustness to outliers (OOD poses).

\vspace{0.02in}
\noindent\textbf{Comparisons.} We compare our method with state-of-the-art learning-based methods that obtain SMPL parameters from point clouds~\cite{Bhatnagar20a, Bhatnagar20b, Wang21}. IP-Net~\cite{Bhatnagar20a} predicts dense correspondences to the SMPL mesh, 
which are then used within an ICP-like optimization that recovers SMPL shape and pose parameters. LoopReg~\cite{Bhatnagar20b} extends this idea by including the SMPL optimization step within the training process of the network. PTF~\cite{Wang21} segments body parts and regresses local transformations for each, from which pose parameters are obtained via least-squares fitting. 
Since all of these methods require part segmentation, we also compare our segmentation results with them. Note that IP-Net and LoopReg predict segmentation for 14 parts, while PTF and our method segment the point cloud into 24 parts, which is a harder task. 
For LoopReg we use their publicly available model trained on minimal clothing data, while the rest of the networks are trained using publicly available code.
All methods, including ours, are trained for 15 epochs on the DFAUST train split. 
Note that all competitors depend on a test-time optimization step, while our results are solely based on regression.

\vspace{0.02in}
\noindent\textbf{SO(3) Data Augmentation.} An alternative to SE(3) equivariant learning is to explicitly perform data augmentation. 
While augmenting articulated body poses is not always feasible
one can easily perform global SO(3) augmentation by randomly rotating the root joint. 
In fact, this is already implemented by prior work, including the methods considered here. For this reason, we compare against these methods both with and without global SO(3) data augmentation. Our method, being an equivariant method, does not require augmentation to achieve good OOD generalization. However, this is still useful since it helps the network bridge the gap between the discretized SO(3) group and the continuous SO(3) group, and hence we too evaluate both with and without SO(3) augmentation.

\subsection{Part Segmentation} 
\label{sec:results_segmentation}

We begin by evaluating our part segmentation network. Both for us and for competing methods, this is an important step in the pipeline that determines the quality of the final results. Since the output of this step comes directly from the networks, initial errors cannot be masked by an optimization step as it is with the SMPL estimations (see next section). We use this task to more clearly evaluate the impact of equivariant learning.

Quantitative results can be found in \Cref{tab:segmentation}. Our method outperforms the competitors both for ID and OOD datasets by a large margin. Without data augmentation, IP-Net and particularly PTF perform very poorly in both cases, 
suggesting a fundamental limitation with respect to generalization of these methods.
Our approach, on the other hand, shows superior performance over all methods both with and without data augmentation. Additionally, we observe that conditioning on the parent feature (\cref{sssec:method_pose_estimation}) can further boost the segmentation accuracy. 
We show qualitative results for our method in \Cref{fig:qualitative_seg} and qualitative results for competing methods in the supplementary material.

\input{figures/tex/ood_results}

\input{tables/tab_comparisons}

\subsection{SMPL Shape and Pose Estimation} 
\label{sec:results_smpl}

In \Cref{tab:comparisons} we show quantitative results for the SMPL body estimation task, evaluated in terms of vertex error and joint position error. Note that our results are obtained directly from the network, while the other methods require an optimization step. For OOD data, our model performs significantly better than the rest, reducing the vertex error by almost two times over the best-performing competitor (PTF with augmentation). This shows the importance of including the right symmetries within the network design, which cannot be replaced by simply performing data augmentation. 
It is worth noting that all the other methods require data augmentation to perform reasonably in the OOD case. While data augmentation slightly improves the results in our case, we outperform the rest even without it. 
For in-distribution data, \modelname without data augmentation also surpasses the previous state of the art model by 68\% in V2V error,
while the competing methods are much slower at inference time due to the optimization step. 
In the bottom part of \Cref{tab:comparisons} we show that our proposed parent conditioning consistently improves the results by around 5 mm for OOD samples.
Qualitative results can be found in \Cref{fig:qualitative_ood}. %

Additionally, we evaluate our model directly on raw real-world scans from the DFAUST dataset \cite{Bogo17}, which contains a different noise distribution including sensor noise and holes. For this OOD noise experiment, we perform inference on these scans \emph{without fine-tuning} our model, and find the part segmentation performance only drops by 5\%. 
While the OOD noise increases the V2V error by 2cm, it is still on par with the accuracy of previous state-of-the-art methods  on  {\em synthetic data}. 
While ArtEq works well without fine-tuning, we suspect that training on real noise will produce significant improvements in robustness and accuracy.
See supplemental material for more information.

\input{tables/tab_times}
\input{figures/tex/convergence}

\subsection{Performance and Convergence}
\label{sec:results_performance}
 
\paragraph{Performance} \Cref{tab:time} shows computation time for all methods, along with model size. Using SE(3)- equivariant/invariant features allows our model to be more space- and time-efficient.
ArtEq has only $2.7\%$ the number of parameters of PTF, while still outperforming in terms of accuracy. Additionally, our smaller model size results in significantly faster inference time, which is three orders of magnitude faster than the others.
Of course, like the other methods, one could add an additional optimization step to refine our results further.

\paragraph{Convergence} In addition to being computationally faster and having lower memory requirements, %
we show in \Cref{fig:convergence} that our method also converges rapidly, requiring merely 5 epochs of training to reach reasonable results, for both in-distribution and out-of-distribution datasets.

\subsection{Limitations}
Our method's primary  failure mode results from  self-contact, a disadvantage shared with most mesh registration methods. 
Here, self-contact may lead to incorrect feature aggregation, where features that belong to different body parts are convolved together due to their proximity in Euclidean space.
Additionally, while our method generalizes to unseen rigid transformations of the body parts, the model is not guaranteed to be robust to {\em non-rigid deformations} that are encoded in the pose corrective blendshapes of SMPL. Its performance in this respect depends on the distribution of non-rigid transformations seen during training. Though data augmentation is helpful to mitigate the aforementioned limitations, a more principled solution to address these is left for future work.

%% file: tables/tab_segmentation.tex
\begin{table}[t]
\begin{center}
    \normalsize
  \scalebox{0.9}{
    \setlength{\tabcolsep}{12pt}
    \begin{tabular}{| l| c | c | c | }
    \hline
    Method & Aug. & OOD & ID \\
    \hline
    IP-Net~\cite{Bhatnagar20a} & & 29.0 & 30.5 \\
    \hline
    IP-Net~\cite{Bhatnagar20a} & $\checkmark$ & 86.7 & 91.2\\
    \hline
    LoopReg~\cite{Bhatnagar20b} & $\checkmark$ & 60.6 & 66.1 \\
    \hline
    PTF~\cite{Wang21} & & 8.5 & 10.3 \\
    \hline
    PTF~\cite{Wang21} & $\checkmark$ & 80.3 & 88.1 \\
    \hline
    \hline
    Ours (nc) & & 91.7 & \underline{96.2} \\
    \hline
    Ours (nc) & $\checkmark$ & \underline{93.8} & \underline{96.2} \\
    \hline
    Ours & & 92.6 & \textbf{96.3} \\
    \hline
    Ours & $\checkmark$ & \textbf{94.1} & \underline{96.2} \\
    \hline
    \end{tabular}
    }
  \vspace{-1mm}
 \caption{Part segmentation accuracy compared to SOTA methods, in terms of percentage of correct predictions, for out-of-distribution (OOD) and in-distribution (ID) datasets. We show results with and without SO(3) data augmentation (``Aug.''), and we show our model with and without parent feature conditioning (``nc'') (\cref{sssec:method_pose_estimation}). Best result in bold, second best underlined.} %
 \label{tab:segmentation}
\end{center}
\vspace{-3mm}
\end{table}

%% file: figures/tex/segmentation_results.tex
\begin{figure*}[t]
\centerline{
\includegraphics[width=0.99\textwidth]{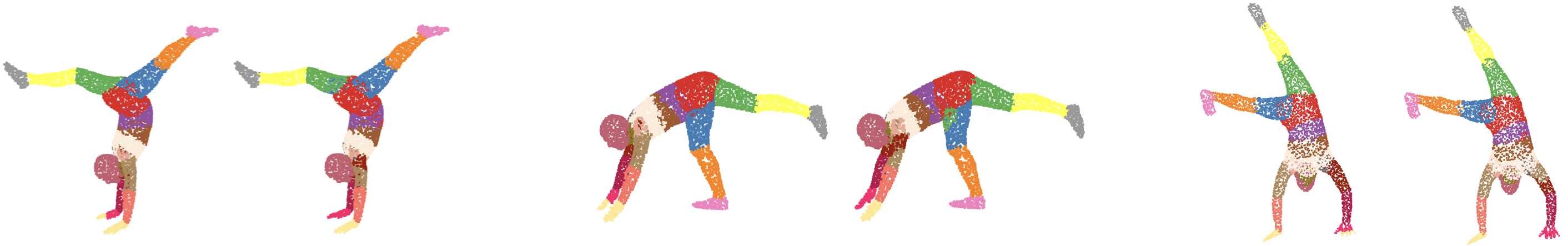}
}
\vspace{-0.075in}
\caption{Qualitative results for part segmentation. Each pair of bodies shows ground-truth  (left) and our result (right). } 
\label{fig:qualitative_seg}
\vspace{-1mm}

\end{figure*}

%% file: figures/tex/ood_results.tex
\begin{figure*}[t]
\centerline{\includegraphics[width=\linewidth]{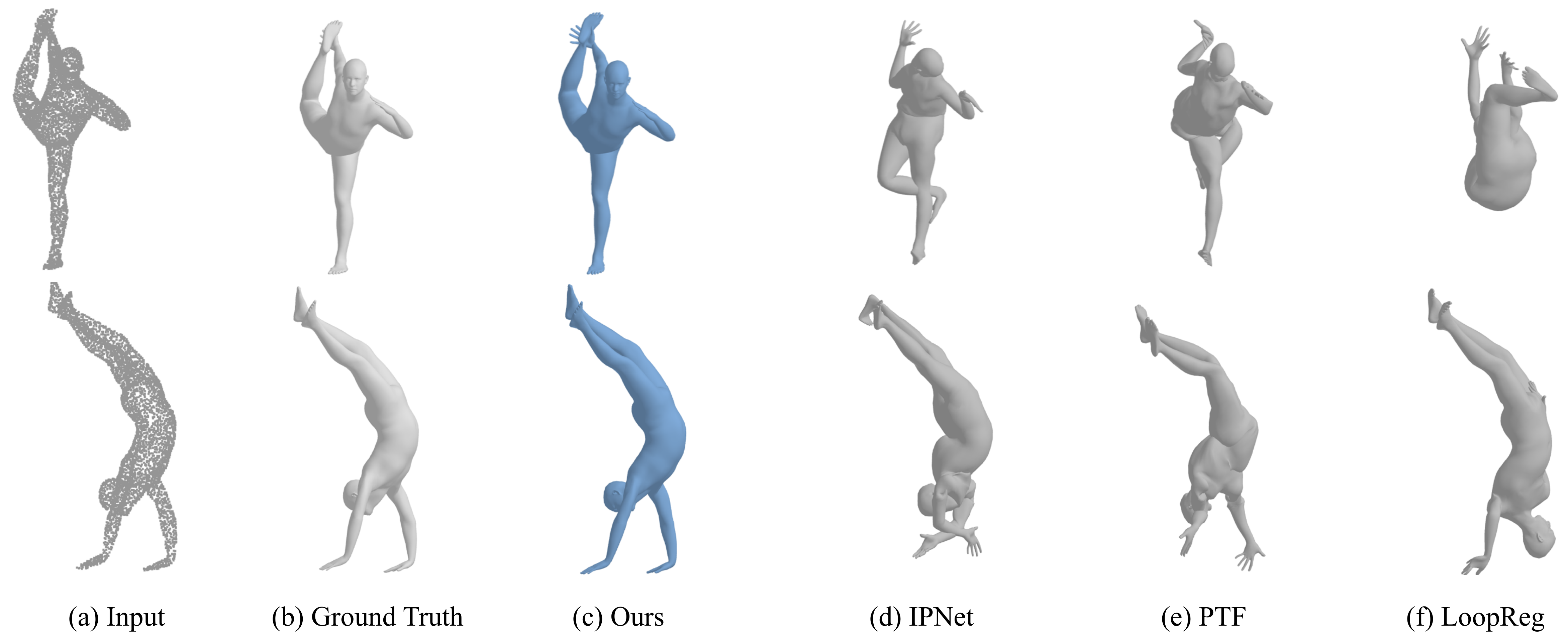}}
\vspace{-0.1in}
\caption{Qualitative results for out-of-distribution poses. From left to right: (a) input point cloud, (b) ground-truth SMPL mesh, (c) our results, (d) IP-Net~\cite{Bhatnagar20a}, (e) PTF~\cite{Wang21}, and (f) LoopReg~\cite{Bhatnagar20b}.} 
\label{fig:qualitative_ood}
\end{figure*}

%% file: tables/tab_comparisons.tex
\begin{table}[t]
\begin{center}
    \normalsize
  \scalebox{0.9}{
    \begin{tabular}{| l|c|cc | cc  | }
    \hline
    Method & Aug.\ & \multicolumn{2}{c|}{OOD} & \multicolumn{2}{c|}{ID}\\
    \hline
     & & V2V $\downarrow$ & MPJPE $\downarrow$    & V2V $\downarrow$ & MPJPE $\downarrow$  \\
    \hline
    IP-Net & & 41.58 & 46.99    & 38.55 & 43.41 \\
    \hline
    IP-Net & $\checkmark$ & 7.57 & 9.41    & 5.98 & 6.42 \\\hline
    LoopReg & $\checkmark$ & 29.08 & 34.09    & 7.57 & 9.17 \\
    \hline
    PTF & & 61.42 & 68.43    & 56.35 & 60.98 \\
    \hline
    PTF  & $\checkmark$ & 6.42 & 7.56   & 3.05 & 3.53 \\
    \hline
    \hline
    Ours (nc) &  & 5.41 & 5.91    & \textbf{0.95} & \underline{1.04}\\
    \hline
    Ours (nc) & $\checkmark$ & \underline{4.17} & \underline{4.61}    & \textbf{0.95} & \textbf{1.03} \\
    \hline
    Ours & & 4.73 & 5.51    & 1.13 & 1.48 \\
    \hline
    Ours & $\checkmark$ & \textbf{3.62} & \textbf{4.23}    & 0.98 & 1.26\\
    \hline
    \end{tabular}
    }
 \vspace{-1mm}
 \caption{SMPL estimation results compared to state-of-the-art methods, with and without SO(3) augmentation (``Aug.'') for out-of-distribution (OOD) and in-distribution (ID) datasets. Metrics: vertex-to-vertex error (v2v, in cm) and mean joint position error (MPJPE, in cm). We also show our method without parent feature conditioning (``nc''). Best result in bold, second best underlined. }
 \label{tab:comparisons}
\end{center}
\vspace{-3mm}
\end{table}

%% file: tables/tab_times.tex
\begin{table}[t]
\begin{center}
    \normalsize
   \scalebox{0.9}{
    \setlength{\tabcolsep}{15pt}
    \begin{tabular}{| l|c | c |}
    \hline
    Method & \#Param (M) & Time (s) \\
    \hline
    IP-Net~\cite{Bhatnagar20a} & 35.0 & 211.4 \\
    \hline
    LoopReg~\cite{Bhatnagar20b} & 3.3  & 146.9 \\
    \hline
    PTF~\cite{Wang21} & 34.1 & 158.1\\
    \hline
    Ours & \textbf{0.9}  &\textbf{0.1} \\
    \hline
    \end{tabular}
    }
 \caption{Number of parameters (\#Param) and inference time (Time) for the methods evaluated in this paper.}
 \label{tab:time}
\end{center}
\vspace{-6mm}
\end{table}

%% file: figures/tex/convergence.tex
\begin{figure*}[t]
\centering
\begin{subfigure}[b]{0.45\textwidth}
\includegraphics[width=\linewidth]{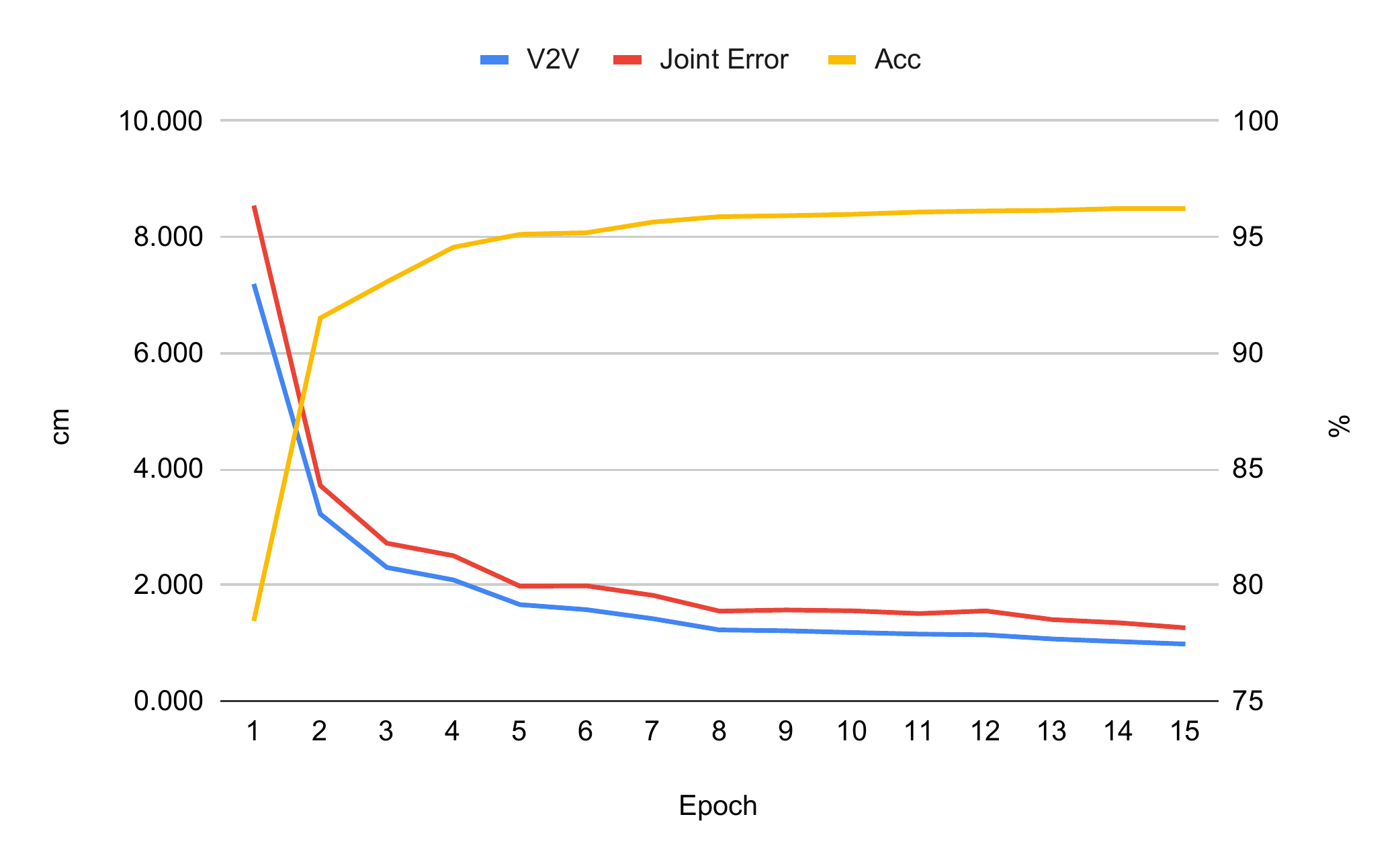}
\caption{In-distribution}
\end{subfigure}
\begin{subfigure}[b]{0.45\textwidth}
\includegraphics[width=\linewidth]{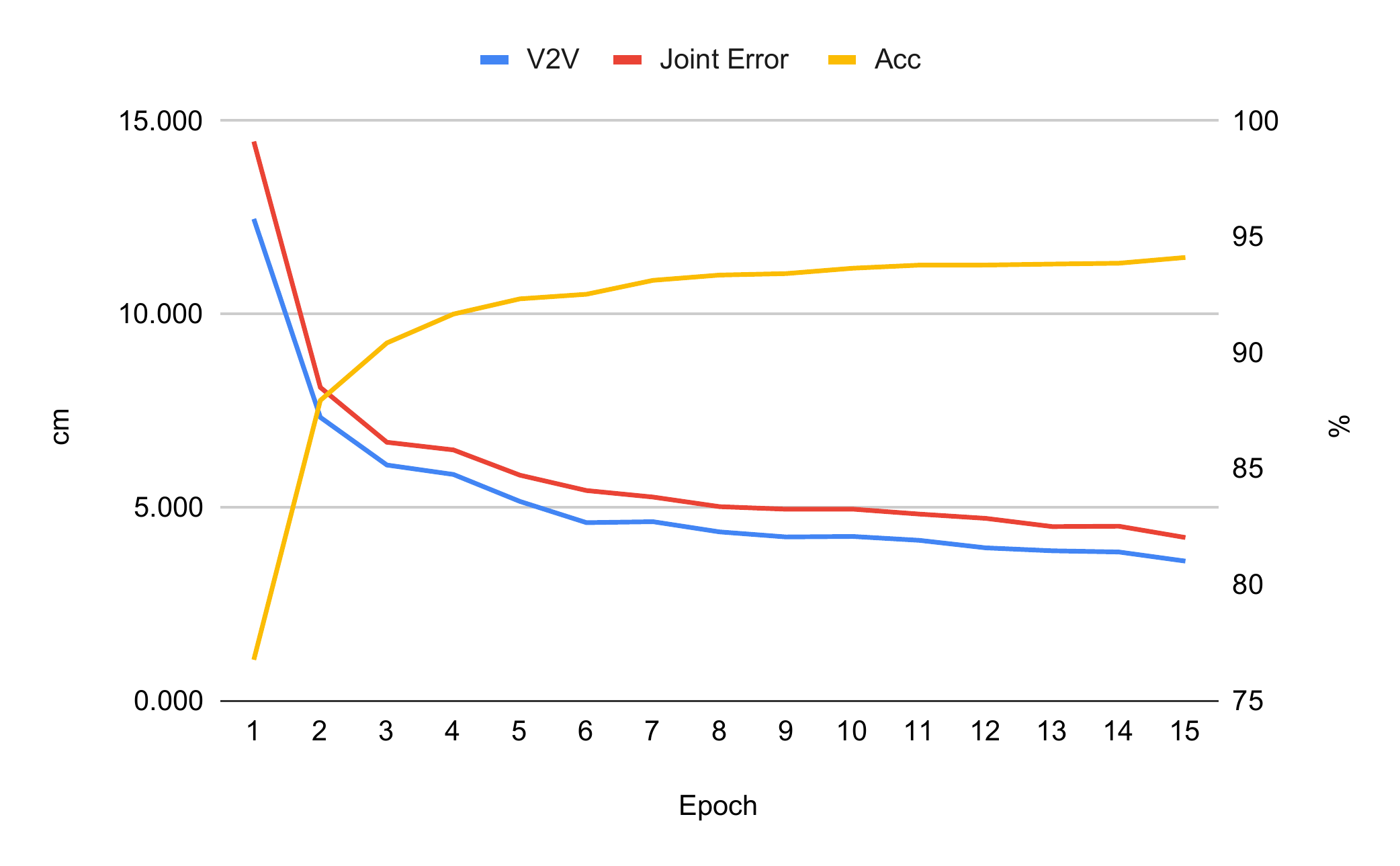}
\caption{Out-of-distribution}
\end{subfigure}
\vspace{-0.1in}
\caption{Convergence plots. Vertex-to-vertex error (\textit{V2V}), joint position error (\textit{Joint Error}), and part segmentation accuracy (\textit{Acc}) as a function of epoch number, on (a) in-distribution and (b) out-of-distribution datasets.}
\label{fig:convergence}
\vspace{-2mm}

\end{figure*}

%% file: sections/5_Conclusions.tex
\section{Conclusions}
In this paper we propose \modelname, a powerful part-based SE(3)-equivariant neural framework for SMPL parameter estimation from point clouds. Our experimental results demonstrate the generalization ability of \modelname to out-of-distribution poses, where the direct regression output of \modelname outperforms state-of-the-art methods that require a time-consuming test-time optimization step. 
\modelname is also significantly more efficient in terms of model parameters and computation.
Our results demonstrate the advantage and importance of incorporating the correct symmetries into the task of SMPL-body pose and shape estimation.
\modelname provides the first fast, practical, method to  infer SMPL models directly from point clouds.
This serves as a foundation to make 3D human mesh registration more accessible and efficient.

\noindent{\bf Acknowledgements.}
We thank Yandong Wen, Nikos Athanasiou, Justus Thies, and Timo Bolkart for proofreading and discussion; Xu Chen, Yuxuan Xue, Yao Feng, Weiyang Liu, Zhen Liu, Yuliang Xiu, Shashank Tripathi, Qianli Ma and Hanqi Zhou for their helpful discussions; and the International Max Planck Research School for Intelligent Systems (IMPRS-IS) for supporting PK.

\paragraph{MJB Disclosure}
\url{https://files.is.tue.mpg.de/black/CoI_ICCV_2023.txt}

%% file: sections/suppmat/results.tex
\section{Additional Results}

In this section we provide additional results, as well as ablation studies that demonstrate the impact of our design choices. 

\paragraph{Qualitative Registration Results} \Cref{fig:sup_qual_ood} shows qualitative results for out-of-distribution testing data, and \Cref{fig:sup_qual_id} shows results for the in-distribution case. IP-Net, PTF, and LoopReg all fail under difficult poses, resulting in unnatural rotations for some body parts. Poses that are particularly far from the distribution, such as standing on the arms, result in very unnatural shapes. In contrast, our method can handle such poses well despite never having seen similar ones during training.
It is worth noting that %
LoopReg uses the same self-supervised objective during training and during optimization, and refines the learned correspondences at test time by overfitting to the input. Hence, we see here that even such a test-time optimization strategy is not sufficient when the initial poses are far from the correct result. 
\input{figures/tex/qual_id}
\input{figures/tex/qual_ood}

\paragraph{Qualitative Segmentation Results} In \Cref{fig:seg_comparisons}
\input{figures/tex/segm_comparisons} we show additional results for part segmentation on out-of-distribution data, along with comparisons to the segmentations obtained by IP-Net, PTF, and LoopReg. Note that IP-Net and LoopReg predict part segmentation for 14 body parts, where, for example, the two shoulder blades, the three spine regions and the hip are all merged into one torso part (here, in red), or the neck is merged into the head region (here, in olive), making it an easier problem. Our method produces accurate segmentations even for these difficult OOD cases, while PTF, IP-Net, and LoopReg struggle to predict the segmentation, particularly in the regions with out-of-distribution pose. For example, in the second row, IP-Net's part segmentation confuses left and right, resulting in a flipped torso with the belly facing up. %

\paragraph{Raw Scan Data} %
We evaluated our method on the raw scans from the DFaust testing set (in-distribution), without any fine-tuning or re-training. Our model  obtains $88.3\%$ accuracy for part segmentation, $3.62$cm vertex-to-vertex error, and $4.37$cm MPJPE error, which is still better than most other methods on clean data.
A qualitative example of these results is shown in \Cref{fig:sup_raw_scans}. Here we see that our estimations are still accurate for out-of-distribution poses, despite the out-of-distribution noise.
\input{figures/tex/raw_scans}

\paragraph{Impact of the Number of Input Points} We show in \Cref{tab:number_of_points} the results of our model when the input is 500, 1000, 2500, and 5000 points. We see here that our method can already perform reasonably well for in-distribution data for 1000 input points, with a segmentation accuracy that is on par with competitors that use 5000 points as input ($91.2\%$ for IP-Net, Table 1 in main paper). The segmentation accuracy does not differ much when moving to the OOD case. The model has lower performance in terms of V2V and MPJPE for a lower number of points on OOD data, however it still outperforms all the competitors (Table 2 in main paper). This shows that our model does not require a significant number of points in order to obtain accurate results, both for in- and out-of-distribution data.
\input{tables/tab_number_of_points}

\paragraph{Baselines Without a Pose Prior} PTF and IP-Net use a pose prior to regularize the pose space when fitting to SMPL. In the main paper we tested these methods with default parameters, which include the use of the pose prior. To make sure that this does not negatively affect the final outcome, we evaluate PTF and IP-Net without the pose prior. The results are shown in \Cref{tab:no_pose_prior}, where we observe that the pose prior does not have a substantial effect on the output. 
\input{tables/tab_no_pose_prior}

%% file: figures/tex/qual_id.tex
\begin{figure*}[t]
\begin{center}
\includegraphics[width=0.9\linewidth]{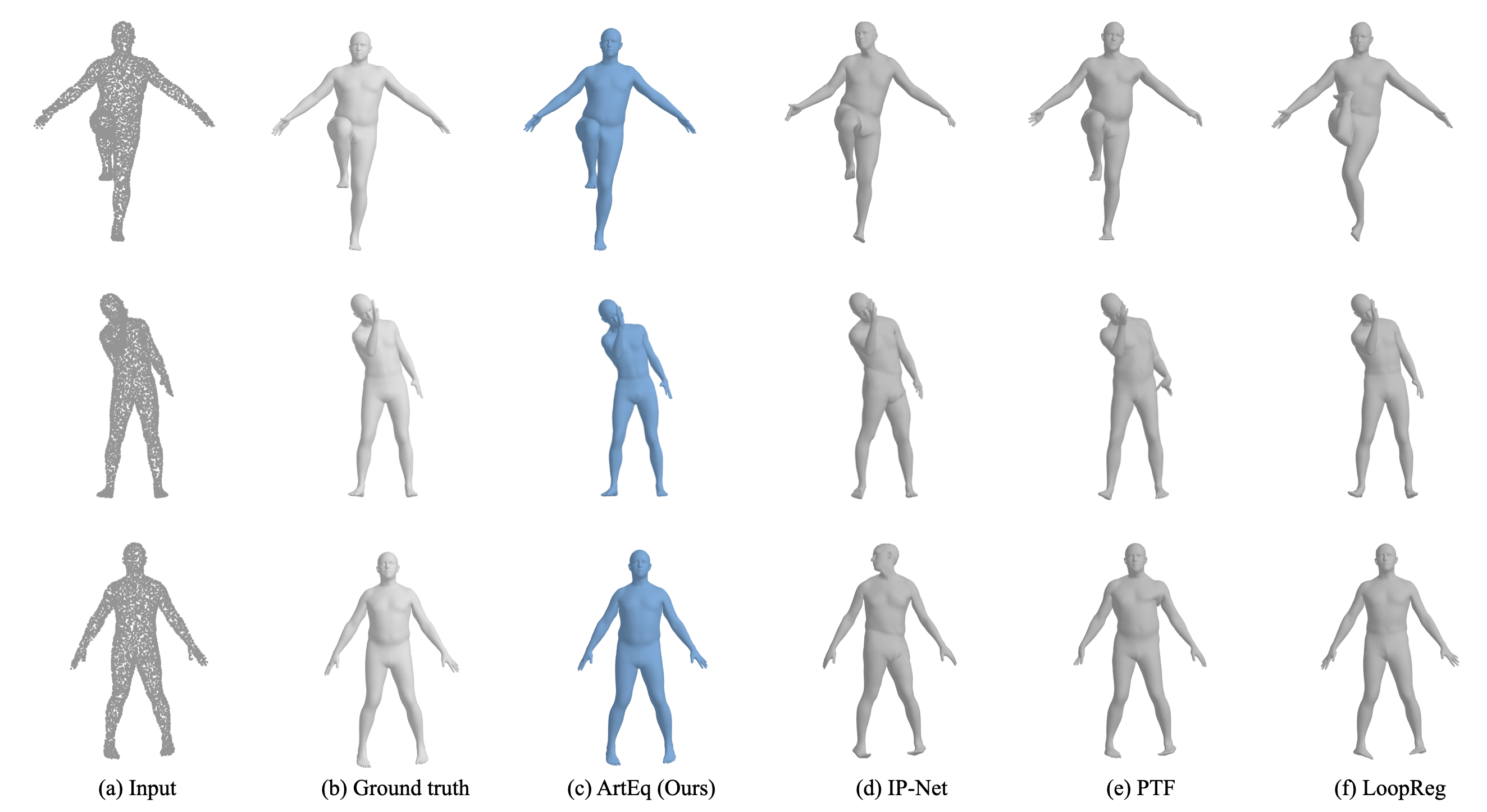}
\vspace{-15pt}
\end{center}
\caption{Qualitative results for in-distribution poses. From left to right: (a) input point cloud, (b) ground-truth SMPL mesh, (c) our results, (d) IP-Net~\cite{Bhatnagar20a}, (e) PTF~\cite{Wang21} and (f) LoopReg~\cite{Bhatnagar20b}.} 
\label{fig:sup_qual_id}
\end{figure*}

%% file: figures/tex/qual_ood.tex
\begin{figure*}[t]
\begin{center}
\includegraphics[width=\linewidth]{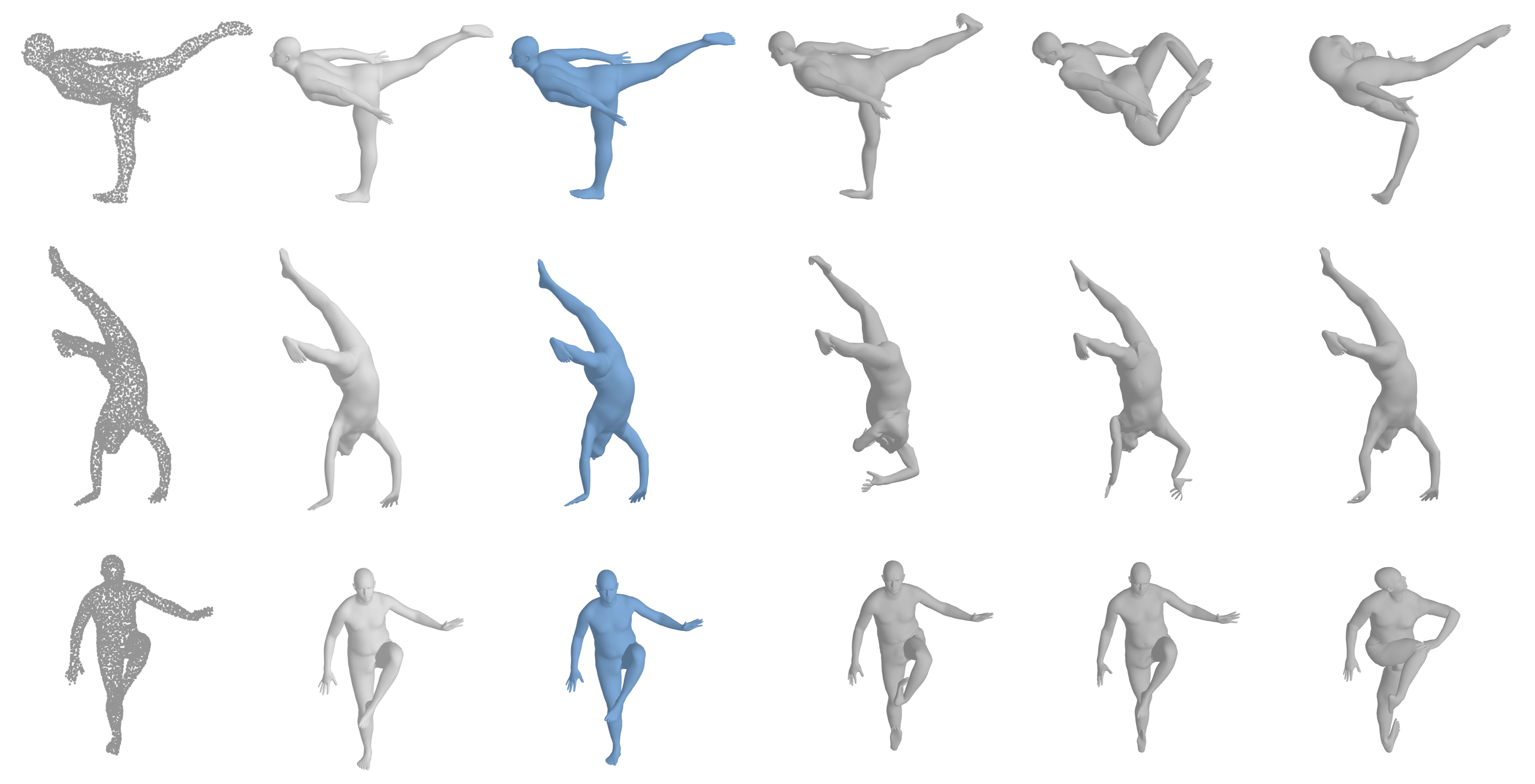}
\includegraphics[width=\linewidth]{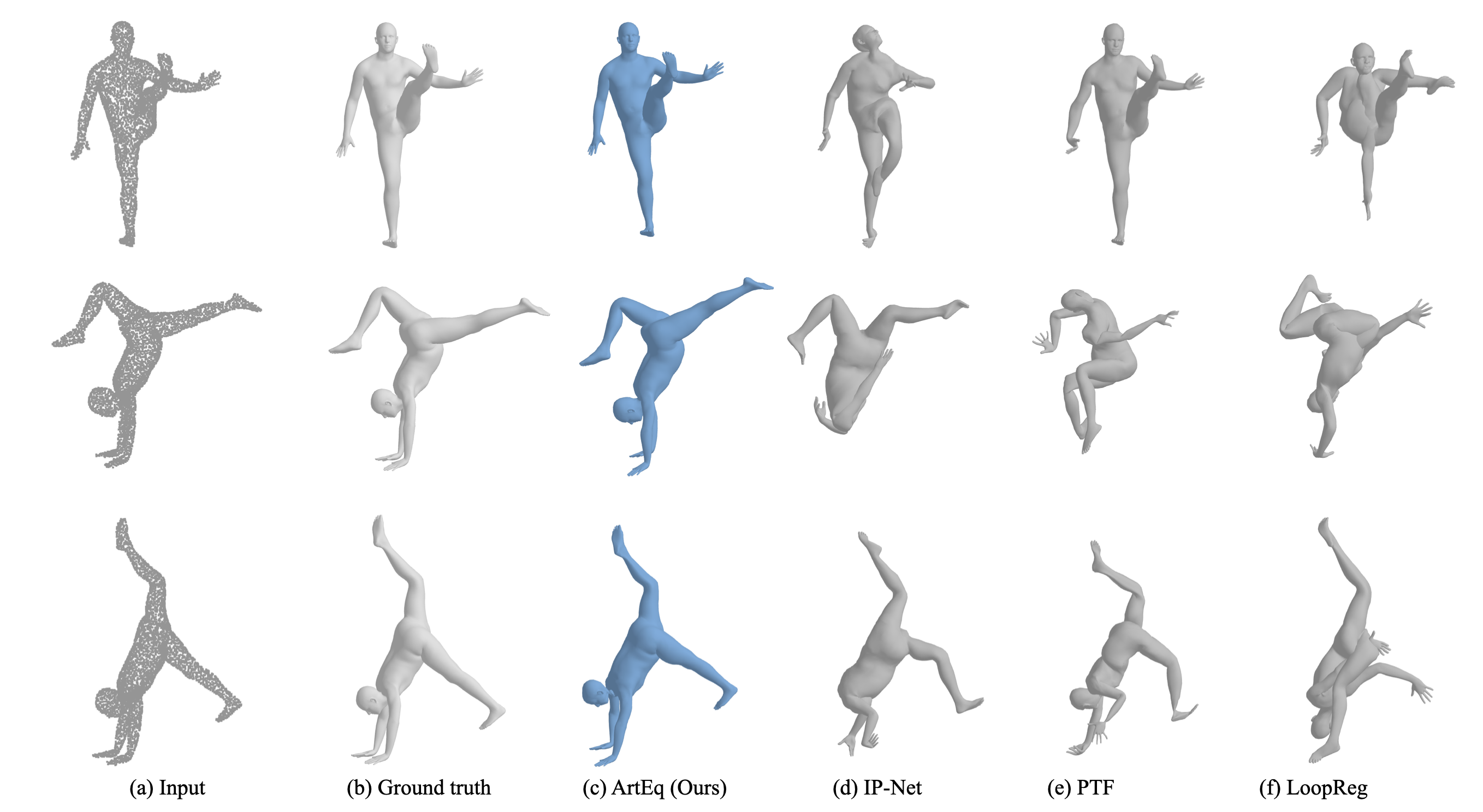}
\end{center}
\caption{Qualitative results for out-of-distribution poses. From left to right: (a) input point cloud, (b) ground-truth SMPL mesh, (c) our results, (d) IP-Net~\cite{Bhatnagar20a}, (e) PTF~\cite{Wang21} and (f) LoopReg~\cite{Bhatnagar20b}.} 
\label{fig:sup_qual_ood}
\end{figure*}

%% file: figures/tex/segm_comparisons.tex
\begin{figure*}[t]
\centerline{
\includegraphics[width=\linewidth]{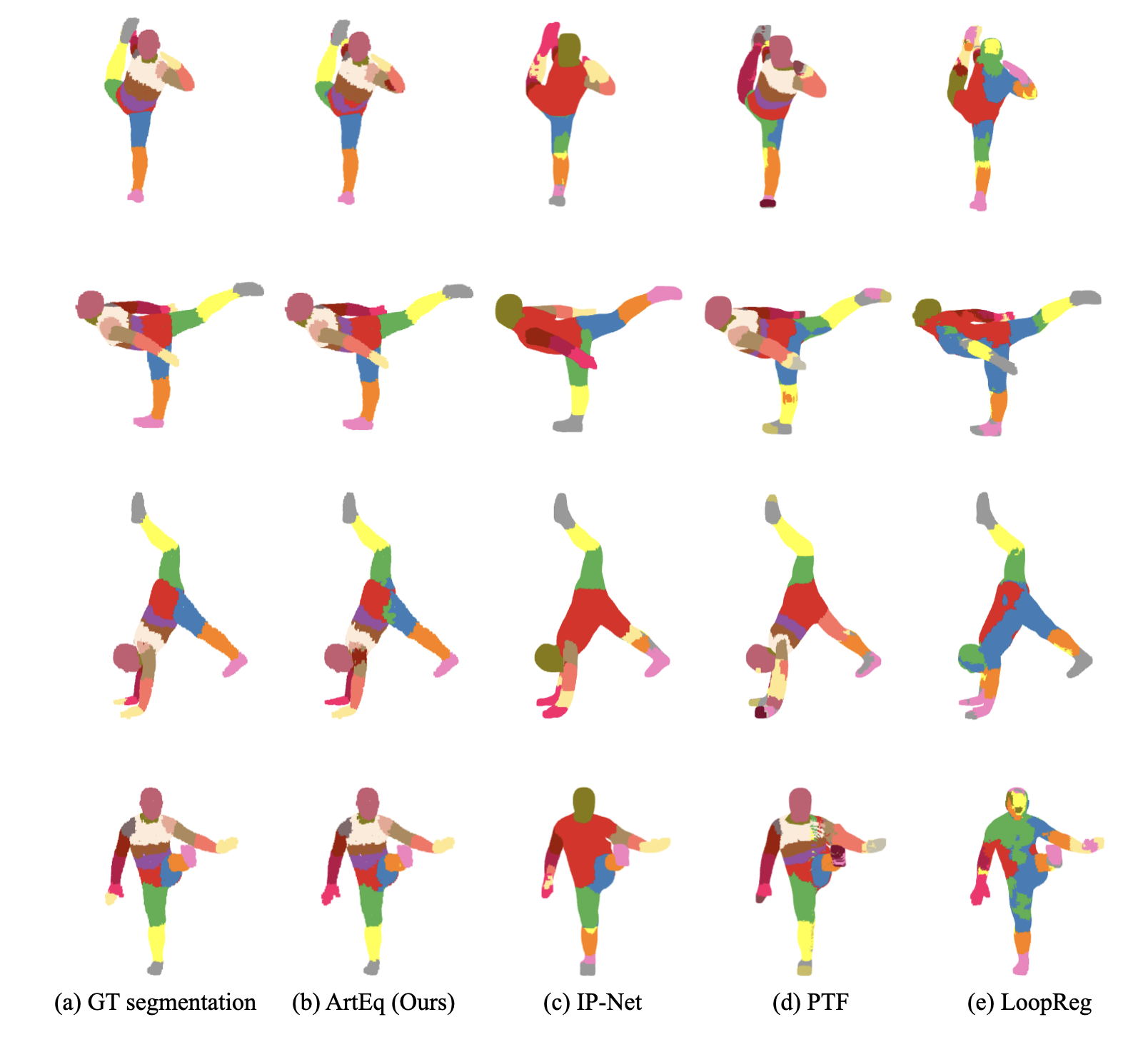}
}
\caption{Qualitative results for part segmentation.  From left to right: (a) ground-truth segmentation, (b) our results, (c) IP-Net~\cite{Bhatnagar20a}, (d) PTF~\cite{Wang21}, and (e) LoopReg~\cite{Bhatnagar20b}.} 
\label{fig:seg_comparisons}
\end{figure*}

%% file: figures/tex/raw_scans.tex
\begin{figure*}[t]
\begin{center}
\includegraphics[width=0.8\linewidth]{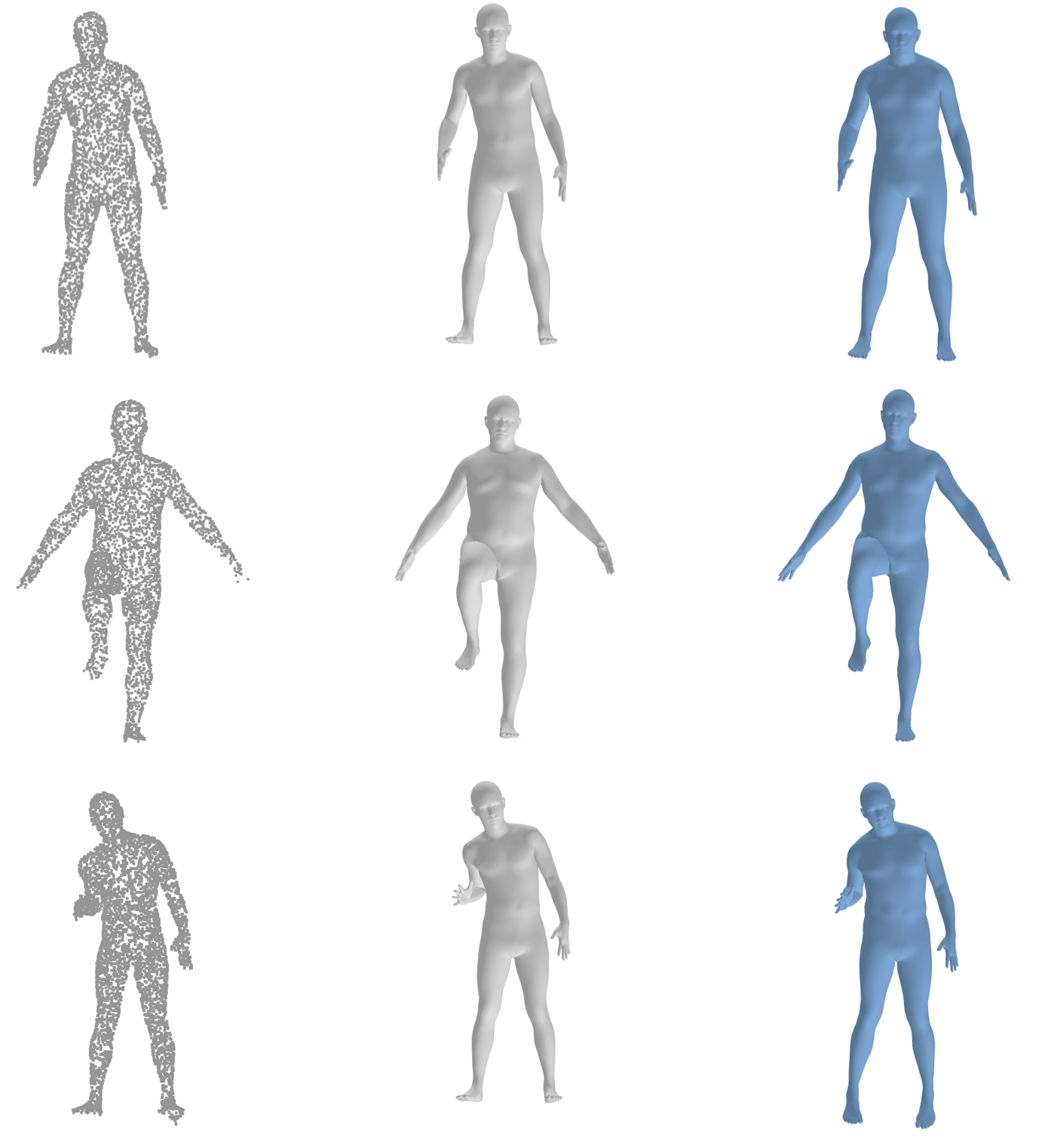}
\end{center}
\caption{Qualitative results on the raw scans from DFAUST testing set. From left to right: (a) input point cloud, (b) ground-truth SMPL mesh, (c) our results.} 
\label{fig:sup_raw_scans}
\end{figure*}

%% file: tables/tab_number_of_points.tex
\begin{table*}
\begin{center}
    \normalsize
    \begin{tabular}{| l| ccc | ccc  | }
    \hline
    \# points & \multicolumn{3}{c|}{OOD} & \multicolumn{3}{c|}{ID}\\
    \hline
    & Seg. $\uparrow$ & V2V $\downarrow$ & MPJPE $\downarrow$    & Seg. & V2V $\downarrow$ & MPJPE $\downarrow$  \\
    \hline
    500     & 80.5 & 7.33 & 8.63    & 82.9 & 4.55 & 5.27\\
    1000    & 89.7 & 4.85 & 5.83    & 92.1 & 2.27 & 2.80\\
    2500    & 93.0 & 4.09 & 4.59    & 95.4 & 1.01 & 1.22\\
    5000    & 94.1 & 3.62 & 4.23    & 96.2 & 0.98 & 1.26\\
    
    \hline
    \end{tabular}
 \caption{Our results for different numbers of input points, in terms of segmentation accuracy (``Seg.''), vertex-to-vertex error (``V2V''), and mean joint position error (``MPJPE'').}
 \label{tab:number_of_points}
\end{center}
\end{table*}

%% file: tables/tab_no_pose_prior.tex
\begin{table*}
\begin{center}
    \normalsize
    \begin{tabular}{| l|c|cc | cc  | }
    \hline
    Method & Pose Prior \ & \multicolumn{2}{c|}{OOD} & \multicolumn{2}{c|}{ID}\\
    \hline
     & & V2V $\downarrow$ & MPJPE $\downarrow$    & V2V $\downarrow$ & MPJPE $\downarrow$  \\
    \hline
    IP-Net & $\checkmark$  & 7.57 & 9.41    & 5.98 & 6.42 \\
    IP-Net &   & 7.67 & 9.55    & 6.04 & 6.50 \\
    
    PTF  & $\checkmark$ & 6.42 & 7.56   & 3.05 & 3.53\\
    PTF & & 6.46 & 7.62    & 3.13 & 3.66\\
    \hline
    Ours &  & \textbf{3.62} & \textbf{4.23}    & \textbf{0.98} & \textbf{1.26}\\
    \hline
    \end{tabular}
 \caption{Comparison with IP-Net and PTF with and without using their pose prior.}
 \label{tab:no_pose_prior}
\end{center}
\end{table*}

%% file: sections/suppmat/implementation.tex
\section{Permutation Equivariance of the Self-Attention Mechanism}
As we have mentioned in the main paper, a function (network) 
$f : V \to W$ is said to be equivariant with respect to 
a group $\group$
if, for any transformation $\mathcal{T} \in \group, f(\mathcal{T} \boldsymbol{X}) = \mathcal{T} f(\boldsymbol{X})$, $\boldsymbol{X} \in V$. Here we elaborate on how the self-attention function $f_{SA}$ is equivariant to the permutation group $\mathcal{T}(\boldsymbol{X}) = \boldsymbol{X} P_\pi$, where $P_\pi$ denotes the permutation matrix of $\pi$, and $\pi$ denotes the permutation of the input tensor's elements (in our case, the permutation over the group element dimension). The self-attention function $f_{SA}$ is defined as $f_{SA}(\mathbf{X}) = \boldsymbol{W}_v \boldsymbol{X} \cdot \operatorname{softmax}\left(\left(\boldsymbol{W}_k \boldsymbol{X}\right)^T \cdot \boldsymbol{W}_q \boldsymbol{X}\right)$, then
\begin{equation}
\begin{aligned}
& f_{SA}\left(\mathcal{T}(\boldsymbol{X})\right) \\
& =\boldsymbol{W}_v \mathcal{T}(\boldsymbol{X}) \cdot \operatorname{softmax}\left(\left(\boldsymbol{W}_k \mathcal{T}(\boldsymbol{X})\right)^T \cdot \boldsymbol{W}_q \mathcal{T}(\boldsymbol{X})\right) \\
& =\boldsymbol{W}_v \boldsymbol{X} P_\pi \cdot \operatorname{softmax}\left(\left(\boldsymbol{W}_k \boldsymbol{X} P_\pi\right)^T \cdot \boldsymbol{W}_q \boldsymbol{X} P_\pi\right) \\
& =\boldsymbol{W}_v \boldsymbol{X} P_\pi \cdot \operatorname{softmax}\left(P_\pi^T\left(\boldsymbol{W}_k \boldsymbol{X}\right)^T \cdot \boldsymbol{W}_q \boldsymbol{X} P_\pi\right) \\
& =\boldsymbol{W}_v \boldsymbol{X}\left(P_\pi P_\pi^T\right) \cdot \operatorname{softmax}\left(\left(\boldsymbol{W}_k \boldsymbol{X}\right)^T \cdot \boldsymbol{W}_q \boldsymbol{X}\right) P_\pi \\
& =\boldsymbol{W}_v \boldsymbol{X} \cdot \operatorname{softmax}\left(\left(\boldsymbol{W}_k \boldsymbol{X}\right)^T \cdot \boldsymbol{W}_q \boldsymbol{X}\right) P_\pi \\
& =\mathcal{T}\left(f_{SA}(\boldsymbol{X})\right),
\end{aligned}
\end{equation}
where we used the property $\operatorname{softmax}(P \cdot A \cdot P^T) = P \cdot \operatorname{softmax}(A) \cdot P^T$, for a permutation matrix $P$ and an arbitrary matrix $A$, to go from the third to the fourth line (refer to the proof in \cite{Yang19}).
Hence, we have proved that the self-attention function is equivariant to the permutation operation over the discretized SO(3) group elements.

\section{Method Details}

\paragraph{Architecture} The local SO(3) feature extractor has two SPConv layers and a nearest neighbor feature propagation layer~\cite{Qi17b}. Each SPConv layer has a kernel size of 0.4 and a stride downsampling factor of 2, therefore, the input point cloud with shape [B, N, 3] will be processed as [B, N/4, 64, 60] where the last dimension is the group element obtained by SO(3) discretization, and $C=64$ is the feature dimension. For each input point, the feature propagation layer finds the top 3 spatial nearest neighbors of the downsampled point-wise features, and interpolates these features weighed by their pairwise distance, resulting in an output of size [B, N, 64, 60].

To obtain the chordal mean weights we attach to the self-attention layers an element-wise MLP (3 layers with ReLU, sizes [64,64,1]), since self-attention does not contain non-linear activations. 
Similarly, we attach a 2-layer MLP on the flattened part features [B, 20*6] to obtain the final SMPL shape code.

\paragraph{Part Segmentation} We consider here 20 body parts, merging the fingers into hands, and toes into feet. This is because the AMASS DFAUST dataset does not contain finger or toe motion.

\paragraph{Averaging Rotations by Calculating the Chordal L2 Mean} Given two rotations $R$ and $S$, the chordal L2 distance is defined as $d_{chord}(R, S) = \lVert R - S \rVert_F$ where $\lVert \cdot \rVert_F$ is the Frobenius norm of the matrix, which is related to the angular distance between $R$ and $S$~\cite{Hartley13}. The chordal L2 mean of a set of rotations is then defined as the matrix that minimizes the chordal distance to all rotations in the set.
In our case, if $\vect{w}_{k, j}$ is the weight for part $k$ and group $j$, then the weighted average for part $k$ over the $|\group|=60$ rotation symmetries is
\begin{equation}
    \argmin_{\hat{\mathcal{R}}_k \in SO(3)} \sum_{j=1}^{\lvert \group \rvert} d_{chord}(\vect{w}_{k, j} \cdot \mathcal{R}(\vect{g}_j), \hat{\mathcal{R}}_k)
\end{equation}
where $\mathcal{R}(\vect{g}_j)$ is the rotation matrix of $\vect{g}_j$, and $\vect{g}_j$ is a group element. In practice, $\hat{\mathcal{R}}_k$ can be obtained in closed-form by using singular value decomposition. We refer the readers to \cite{Hartley13} for more details.

\paragraph{Loss Function} We train both stages of the network with the following loss function:
\begin{equation}
    \lambda_1 \mathcal{L}_{pose} + \lambda_2 \mathcal{L}_{shape} + \lambda_3 \mathcal{L}_{verts} + \lambda_4 \mathcal{L}_{joint} + \lambda_5 \mathcal{L}_{part},
\end{equation}
where 
\begin{itemize}
    \item $\mathcal{L}_{pose} = || \tilde\posecoeff - \posecoeff ||^2 $ is the MSE loss between predicted pose coefficients $\tilde\posecoeff$ and ground-truth pose coefficients $\posecoeff$.
    \item $\mathcal{L}_{shape} = || \tilde\shapecoeff - \shapecoeff ||^2 $ is the MSE loss between predicted shape coefficients $\tilde\shapecoeff$ and ground-truth shape coefficients $\shapecoeff$.
    \item $\mathcal{L}_{verts} = || \mathcal{W} \left( M(\tilde\shapecoeff, \tilde\posecoeff) - M(\shapecoeff, \posecoeff) \right) ||^2 $ is the weighted MSE loss between the reconstructed SMPL mesh vertices and the ground-truth registration, using the per-vertex weights $\mathcal{W}$, where the vertices corresponding to body markers are assigned a weight of 2.0, and the other vertices a weight of 1.0.
    \item $\mathcal{L}_{joint} = || \mathcal{T} ( \mathcal{J}(\tilde\shapecoeff), \tilde\posecoeff)  - \mathcal{T} ( \mathcal{J}(\shapecoeff), \posecoeff) ||^2 $ is the MSE loss between the predicted joint positions of the SMPL mesh (posed) and the ground-truth joint positions.
    \item $\mathcal{L}_{part} = \operatorname{cross-entropy}(\alpha(\point, \partk), \alpha_{gt}(\point, \partk) ) $ is the cross-entropy loss between the predicted part segmentation and the ground-truth part segmentation of the point cloud.
\end{itemize}
We use $\lambda_1 = 5$, $\lambda_2 = 50$, $\lambda_3 = 100$, $\lambda_4 = 100$, $\lambda_5 = 5$.